# Why robots? A survey on the roles and benefits of social robots in the therapy of children with autism

John-John Cabibihan[1], Hifza Javed[1], Marcelo Ang Jr.[2], and Sharifah Mariam Aljunied[3]

[1]*Department of Electrical and Computer Engineering, National University of Singapore*

[2]*Department of Mechanical Engineering, National University of Singapore*

[3]*Education Services Division, Ministry of Education, Singapore*

# Abstract

This paper reviews the use of socially interactive robots to assist in the therapy of children with autism. The extent to which the robots were successful in helping the children in their social, emotional, and communication deficits was investigated. Child-robot interactions were scrutinized with respect to the different target behaviours that are to be elicited from a child during therapy. These behaviours were thoroughly examined with respect to a child's development needs. Most importantly, experimental data from the surveyed works were extracted and analyzed in terms of the target behaviours and how each robot was used during a therapy session to achieve these behaviours. The study concludes by categorizing the different therapeutic roles that these robots were observed to play, and highlights the important design features that enable them to achieve high levels of effectiveness in autism therapy.

# Introduction

Autism is a developmental disorder that encompasses a large variety of disorders with impairments in social relationships, communication, and imagination, and with the severity and nature of the symptoms varying from one individual to another. Autism is a life-long disorder for which no cure has yet been found. With early intervention, much can be done to improve the quality life of those who are afflicted. Several therapeutic approaches have been attempted over the years. However, due to the nature of the disorder and the large variations in the symptoms, no single approach can be established as the best one since the therapy model that may work well with one child may not work well at all with another.

In this survey, we present the emerging works on social robots in the therapy of children with autism. We first give an overview of autism and its rate of occurrence in many countries in the world. Next, we describe each robot's features and their respective effects on a child with autism. We then highlight the behaviours that the robots were tasked to evoke from a child. Finally, we discuss the roles and the therapeutic benefits of social robots for children with autism. The present account was intended to provide introductory-level information for robot designers as well as to familiarize clinicians and parents of children with autism with the recent developments in robot technologies and how these can be helpful in therapy.

*Autism's triad of impairments*

The core impairments in Autism Spectrum Disorder (ASD) are often described along three key dimensions [1,2]:

- **Social relationships/interaction**: These difficulties range from complete indifference

towards other people to a desperate need to make friends, but not being able to do so due to the inability to understand social cues, others' behaviours, and feelings [3,4]. There is an absence of the mentalizing ability (or the 'Theory of mind' [5]) in many individuals with autism, that is the ability to form an understanding about what other people are thinking or feeling. Social situations are extremely challenging and result in the individual wanting to avoid interaction altogether. Abilities to make eye contact, to interpret feelings, to understand the tone of voice, and to read facial expressions all lack in an individual with autism, all of which affect the normal development of social relationships. As a result, children with autism often find difficulty in cooperative play with other children. Instead, they prefer to continue with their own repetitive activities that eliminate participation of others [6]. Impairments in social and environmental exploration prevent these children from learning many fundamental skills and hinder their developmental progress. They may also show a lack of interest in physical interaction, have discomfort while making eye contact, and lack emotional sensitivity towards other children's reactions to them [6].

- **Social communication**: Difficulties in this area include both verbal and non-verbal communication [4]. Speech may be completely absent in some cases, or may be present but with impairments in tone and pitch variation. This gives rise to impairments in the use of intonation and also a lack of understanding of other people's use of it [3]. Speech, when present, may be repetitive and focused on the individual's own obsessive ideas rather than having relevance to the conversation. This is a direct consequence of the inability to read into the deeper meaning of what is being said or done. This impairment also manifests in the form of the inability to initiate and contribute to conversations. For all individuals, typically developing or those with autism, non-verbal communication, such as facial expressions, body language and social cues, are not innate but are learnt with time from one's social environment [3]. Therefore, an individual with autism can exhibit a limited understanding and expression of emotions, gestures, body language, and other cues. These have serious consequences in their social communication abilities.

- **Imagination**: Deficits in imaginative and conceptual skills sometimes lead to the inability to generalize the skills that have been learnt in isolation and to think in abstract terms. These result in a rigid way of thinking and doing things, repetitive activity, and narrow interests [4]. For children with autism, changes in routines are often met with anxiety and distress [3]. These changes could encompass variations in play patterns, food choices, activity schedules, and other daily activities. Their preference for rigid patterns and routines in activities may be a reflection of their attempts to make sense of the world around them [6]. Alongside this rigidity, there could also be a lack of imagination and play skills, and difficulties in incidental learning. This has the most profound impact on the daily lives of these children and their families, since it strongly affects how these children are managed at home and in schools [6].

The impairments and the typical difficulties encountered by the children with autism are summarized in Figure 1.

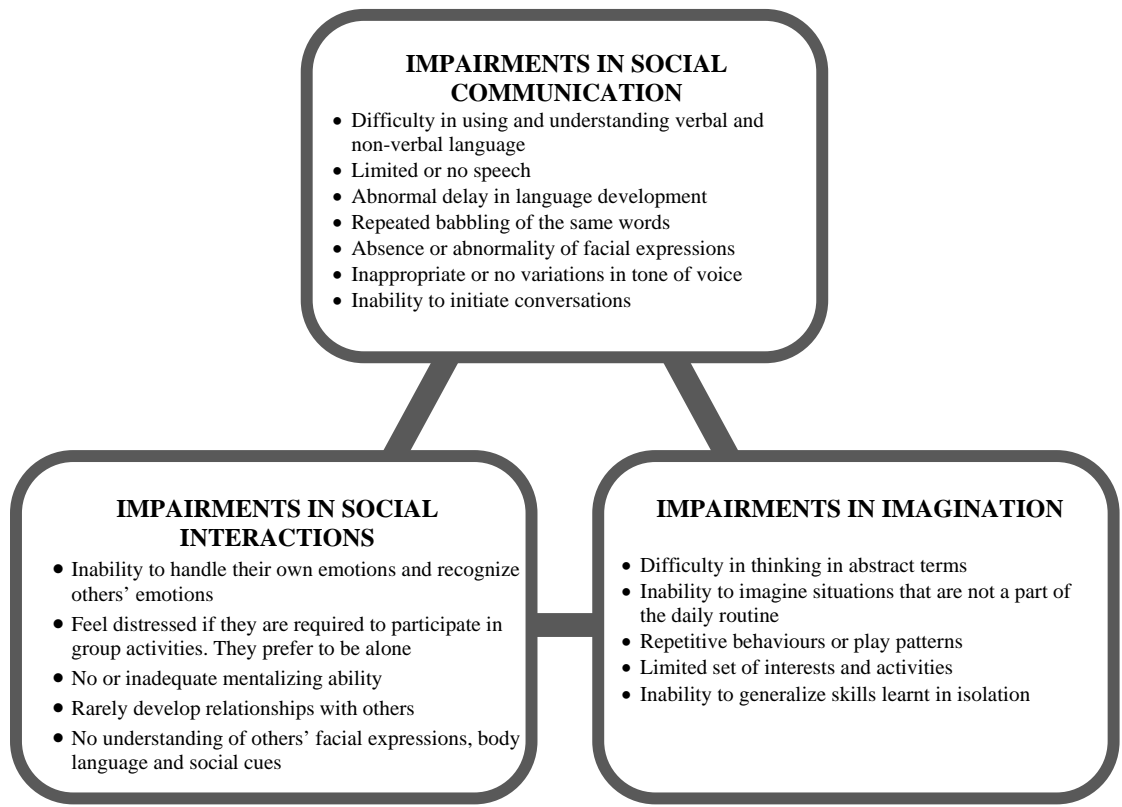

Figure 1. The triad of impairments in Autism Spectrum Disorder (ASD)

According to the current classification scheme in DSM-IV [1], the umbrella term Pervasive Developmental Disorders (PDD)[1] has been used to describe a wide spectrum of disorders, including autism and Asperger's syndrome [8].

Asperger's Syndrome is at the higher functioning end of the autism spectrum. Individuals with this syndrome have to learn specific rules about what kind of behaviour is appropriate in order to succeed socially [9,10]. Instead of naturally developing the ability to interpret social cues, they often memorize rules that govern their social behaviour in order for them to live independently and thrive professionally. They are capable of functioning well in everyday lives but they may face difficulties in social interactions [4]. They often have a limited set of interests and their habits may appear to be bizarre to others. They may also be very clumsy and show motor delays. Boys are approximately three to four times more likely than girls to be affected by Asperger's Syndrome [11]. Children suffering from this syndrome do not have impaired intelligence and often perform from average to above average in academics [12].

---

[1] However, in the soon to be published revised diagnostic criteria described in DSM-V [7], the term ASD is expected to replace PDD. Until then, both the terms are used and understood to mean the same.

It must be mentioned that not every child with autism or Asperger's Syndrome shows all these impairments to the same degree; each exhibits a subset of the symptoms with varying intensities [6,13]. According to DSM-IV, for a child to be diagnosed with ASD, the child must show at least 6 of the symptoms across the categories in the impairment triad [8,14], with at least 2 symptoms from *impairments in social interactions*, and one each from *impairments in social communication* and *impairments in imagination* [1]. A child can be diagnosed with autism by the age of 3 only after the child has exhibited all of the three impairments. The diagnosis and assessment of ASD involves a multi-disciplinary team of specialists and clinicians, and the use of specialised clinical interviews and observation techniques [7]. Currently, no single root cause of these impairments has been identified. Some researchers [13] believe that the impairments associated with ASD have multiple, largely independent causes at the genetic, cognitive, and neural levels.

*Rate of autism occurrence*

The Center for Disease Control and Prevention estimated the rate of autism occurrence to be 1 in every 88 children in the United States [15]. In South Korea, the estimate was about 1 in every 38 children [16]. According to Singapore's Autism Resource Centre, the figure in Singapore is estimated to be 24,000 in a population of 4 million [17]. Of these, 5,472 are children under the age of 19 years. In general, it was observed that ASD occurrence is more common among boys than girls (1 in 54 in boys and 1 in 252 in girls), and is extended to all races and ethnicities [18].

Other researchers have attempted to quantify the number of people affected by autism (Table 1; [19]). Some of the previous studies may not have produced completely reliable results due to a variety of reasons: lack of qualified professionals for the diagnosis and poor healthcare facilities being the major ones. Results are also influenced by the differences in approaches used to identify cases of autism [19].

**Table 1** Estimated rates of occurrence of autism in different countries

| Study | City, Country | Site Information | Number of cases per 10,000 people | Ages | Diagnosis |
| --- | --- | --- | --- | --- | --- |
| Al-Farsi Y.M. et al, 2010 [20] | Muscat, Oman | One hospital at Sultan Qaboos University | 1.4 | Up to 14 years | ASD |
| Baron-Cohen S. et al., 2009 [21] | Cambridge, United Kingdom | 96 schools in Cambridgeshire County | 94 | 5 to 9 years | ASD |
| Fombonne E. et al., 2006 [22] | Montreal, Canada | 55 schools in Montreal | 64.9 | 5 to 17 years | PDD |
| Kawamura Y., 2008 [23] | Toyota, Japan | One center in Toyota | 181.1 | 5 to 8 years | PDD |
| Oliveira G et al., 2007 [24] | Lisbon, Portugal | Random sampling from schools in Portugal | 16.7 | 6 to 9 years | ASD |
| Parner E.T. et al., 2011 [25] | Perth, Australia | Western Australia Register for Autism Spectrum Disorders | 51 | Up to 10 years | ASD |
| Parner E.T. et al., 2011 [25] | Copenhagen, Denmark | Danish National Psychiatric Registry | 68.5 | Up to 10 years | ASD |
| Paula C.S. et al., 2011 [26] | Sao Paulo, Brazil | Schools and health services in Atibaia, Sao Paulo | 27.2 | 7 to 12 years | PDD |

*Social robotics and autism*

Socially interactive robots are used to communicate, express and perceive emotions, maintain social relationships, interpret natural cues, and develop social competencies [27,28]. Social robots are being used as tools to teach skills to children with autism, to play with them, and to elicit certain desired behaviours from them. They create interesting, appealing, and meaningful interplay situations that compel children to interact with them. One of the emerging applications of social robotics is the therapy of children with autism [29-32]. Although several interactive software agents and computer-mediated therapy models have been developed for autism therapy to exercise different skills, such as computer-mediated imaginative story telling [33], the research works that will be described in the succeeding sections suggest that socially-interactive robots could perform much better.

In recent past, interest in this field has grown tremendously, leading to research initiatives taken by several agencies and universities across the world to develop robots and conduct clinical tests on children with autism. A multitude of robots have been created, all of which vary in appearance, behaviour, and activities that they are capable of doing. The following sections give a comprehensive study of the available literatures that do not only describe the robot's features and their significance, but also describe the purpose of each feature and the experimental methods employed to achieve them.

# Robot design features

The design and functionalities of the robot have a significant influence on its effectiveness in therapy. Children with ASD may show more receptiveness towards some features but discomfort to others. However, it must be pointed out that although some similarities in children's reactions to certain robot features do exist, due the nature of the disorder, not all children with ASD will react exactly the same way.

To ensure the suitability of the robot's design, several research studies have been conducted to elicit requirements from the actual end-user group, that is, the children with autism. However, since these children themselves have impaired communication, panels composed of experts, therapists, parents, and teachers were asked to give their feedback [34]. Other efforts have also been made to compile a detailed set of design requirements that are not subjective, but can be generalized to most of the children's preferences [35,36]. To get the children's direct perspective, some experimenters asked a large sample of normally developing children ($n$=159) to evaluate 40 robot designs using questionnaires, and analysed their responses to evaluate children's attitudes and feelings towards various attributes [37]. This analysis could be used to distinguish between the design preferences of children with autism from their typically developing peers (i.e. the control group).

From the study of a variety of robots used for autism therapy from the available literature, we categorized the robot design requirements according to appearance, functionality, safety requirements, autonomy, modularity, and adaptability.

*Appearance*

*Visual appeal*

It is vital for robots in therapy to be visually engaging for a child with autism, since these children are known to exhibit short concentration spans [35,36,38]. While brightly-coloured body parts grab attention, they must not be so bright that they over-simulate the child [35,39]. Different body parts can be coloured differently for emphasis. Children also find different shapes, lights, and mechanical rotating parts appealing [35]. However, sharp edges, ropes and bright colours must be avoided.

*Realism*

Interactions with robots are engaging for children with autism because of the reduced complexities as compared with human interactions. Therefore, the robot must not be too human-like or the child may lose interest [38,40]. Complex facial expressions may also be avoided, alongside trivial features such as eyebrows and eyelashes to enhance simplicity. While most of these children are uncomfortable in making eye contact and may feel threatened by the robot's eyes, some seem to be attracted to this feature [41]. As such, these children can be encouraged to initiate and maintain eye contact. However, this must be a modular feature since this preference varies considerably from child to child [38]. A balance must be considered to ensure that the robot is not so human-like that the child feels threatened and not so mechanical that the child gets more interested in examining the robot's mechanical parts [42]. However, it is essential that the child is always aware that the robot is, indeed, a mechanical being and not a human.

*Size*

Therapists have found that the most appropriate size for a therapeutic robot must be roughly the size of the child who is undergoing therapy [43]. Since the targeted user of the robot is a child, making the robot the same size as himself/herself will allow for easier and more enjoyable interaction and play. Eye contact is easier to make since the heights are similar. The skills they learn with such a robot can then be extended to other children. Additionally, a robot that is approximately the size of the child can be less intimidating [36].

*Anthropomorphic, non-anthropomorphic or non-biomimetic*

In the robotic therapy projects that have been undertaken to date, a wide range of robot types has been employed. The robots are either intended to possess an acute resemblance to humans (anthropomorphic), or they are designed as animals or cartoon-like toys (non-anthropomorphic), or they are designed to not resemble any biological species (non-biomimetic). This variation, as summarized in [44], ranges from human-like androids, to cartoon-like mascots, to mechanical humanoids, to animal-like robots, to non-humanoid mobile robots. Children diagnosed with autism tend to avoid interactions with others. Thus, experiments have been undertaken to evaluate the response to humanoid versus non-humanoid robots. Results show that children have a general fondness for non-humanoid designs [40]. Children tend to show greater stimulation in response to robots with pet-like or cartoon-like features [35,45-47]. They also respond positively to non-biomimetic robots such as mobile, vehicular robots [35,38]. On the other hand, humanoid robots may be preferred in some cases because it is felt that imitation and emotional skills taught through these robots are easier to generalize to other humans. Robots with overly mechanized appearances may also not derive the best results since too many exposed mechanical parts can cause the child to shift focus from the interaction itself [45,48].

*Functionality*

*Sensory rewards*

For a child with autism, it is important that the correct execution of a task is encouraged to ensure that the child feels rewarded for the achievement [38]. A task may appear to be trivial to a typical adult or child, but it may take much effort for such a child to execute it as instructed. Hence, explicit positive feedback proves to be highly beneficial. This encouragement can be given in the form of sensory rewards, such as the lighting up of the robot's body part, or the playing of some music, or the robot's clapping [35]. Children find these rewards extremely intriguing and encouraging [34,35].

*Locomotion*

Many robots use locomotion to attract a child's attention. Children with autism tend to be more attracted towards moving things, thus making the robot's ability to move an important factor. Experiments have shown that they prefer to play with interactive, robotic toys rather than passive toys [49]. Mobile robots have been shown to elicit positive behaviours from these children [35,50]. Furthermore, it is an additional advantage if the robot is also capable to moving other objects, such as kicking, throwing a ball or moving blocks [51]. These features allow for enhanced play scenarios to be developed since the robot is able to play an important role in the games [34,48].

*Choice and control*

A child with autism could be taught more effectively if the child has the ability to make choices during the interaction with the robot. For example, if the child could choose between the blinking of an LED or the playing of music from the robot after the child greets it as instructed, it can keep the child more interested in "making the interaction happen" and give the child more control [38]. Control buttons on the robot could allow this feature to be implemented, such as pressing different buttons that leads to different consequences.

*Safety requirements*

Children with autism can be uncontrollably exuberant or impulsive at times. Consequently, they are prone to touch the robot and mishandle it. In doing so, the children may hurt themselves and the robot. It must be ensured that the robots have no sharp edges, do not have, fast, or jerky movements, in addition to minimum probability of malfunctioning. To ensure the robot's safety, the robot must be robust. During a child's meltdown, the designer can explore robot designs that can withstand being dropped on the floor or being thrown to a wall.

*Autonomy*

The robot must be able to exhibit a high level of autonomy such that the need to control its every action is eliminated [52]. It must be able to execute a sequence of desired motions without being controlled by the therapist. One button pressed on the remote should result in a series of steps that reflects a specific kind of behaviour needed to be communicated to the child, instead of a few trivial, insignificant motions [36]. On the contrary, complete autonomy may not be desirable since no robot can replace a human completely; the therapist must be able to decide the robot's behaviour in response to the behaviour of the child. A human may lack the robot's repetitive nature and its ability to exhibit only a small set of emotions, but he/she is still a better judge of how the robot must respond to a child's behavior, instead of letting the robot take complete control of the interaction. Hence, the presence of a human in the loop is vital.

*Modularity and adaptability*

The nature of ASD is such that each child's interests, preferences, and capabilities may significantly vary from the other. Modularity allows different functionalities to be carried out for different children, enabling them to choose amongst robot features that can sustain their interest [38]. On the hardware aspect, the structure of the robot can be made modular whereby, if one part is damaged, there is no need to replace the entire robot [36]. The robot must also have a high level of adaptability to a specific environment or a child. It must be able to show a progressive growth in the complexity of its interactions with the child's development [35]. This progressive growth in interactions, such as games, ensures that the child is continuously trained with new skills and abilities.

## Robots for autism research

Table 2 presents a compilation of the different robots that have been studied in this survey. These robots are used in autism therapy for children in different parts of the world with different levels of success. Each has its unique appearance, design features, and interaction methods. Each is aimed at evoking one or more specific types of behaviours from the children being treated.

Table 2: Compilation of robots employed in autism therapy for children

| | Robot | Figure | Type | Features |
|---|---|---|---|---|
| 1 | Bobus [35] | 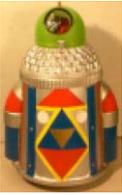<br>(©2003 IEEE. Reprinted, with permission, Characteristics of Mobile Robotic Toys for Children with Pervasive Developmental Disorders, Michaud F, Duquette A, Nadeau I) | Non-anthropomorphic | • The robot can detect a child's presence<br>• Robust<br>• Plays music and performs simple movements when the child is in close proximity<br>• Light emitting diodes (LED) all around the neck |
| 2 | CHARLIE (Child-centred Adaptive Robot for Learning in an Interactive Environment) [53] | 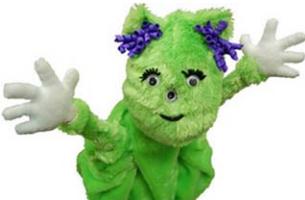<br>(With kind permission from Springer Science+Business Media: Intl J Soc Rob, CHARLIE: An Adaptive Robot Design with Hand and Face Tracking for Autism Therapy, 2011, Buccanfuso L, O'Kane JM) | Anthropomorphic | • The robot has a head and two arms<br>• Camera for face and hand detection<br>• Low cost and simple hardware<br>• Robust and safe<br>• Automatically generates a summary of interactions |
| 3 | CPAC [35] | 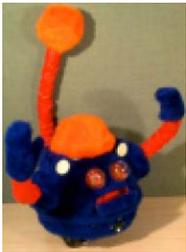<br>(©2003 IEEE. Reprinted, with permission, Characteristics of Mobile Robotic Toys for Children with Pervasive Developmental Disorders, Michaud F, Duquette A, Nadeau I) | Non-anthropomorphic | • Has a tail, arms and LED eyes<br>• Robust and modular, with removable parts<br>• Can rotate on itself<br>• Can dance on a press of a button<br>• Maintains close distance to child using sensors |
| 4 | Diskcat [35] | 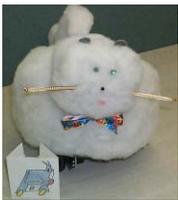<br>(©2003 IEEE. Reprinted, with permission, Characteristics of Mobile Robotic Toys for Children with Pervasive Developmental Disorders, Michaud F, Duquette A, Nadeau I) | Non-anthropomorphic | • Cat-like with fur exterior<br>• Whiskers made of resistive bend sensors<br>• Can play games like "Simon says"<br>• LEDs in place of eyes and at the back for visual appeal<br>• Capable of dancing |
| 5 | FACE (Facial Automation for Conveying Emotions) [54] | 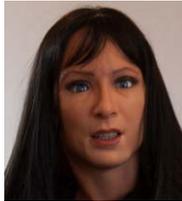<br>(©2010 IEEE. Reprinted, with permission, The FACE of Autism, | Anthropomorphic | • A female android<br>• The face is made of skin-like silicone rubber<br>• Motors move artificial skin on face<br>• Limited set of facial expressions: 6 basic emotions (happiness, sadness, surprise, anger, disgust, fear)<br>• Face and eye tracking through camera |
| | | | | Mazzei D, Billeci L, Armato A, et al) |
| 6 | HOAP-2 [55] | 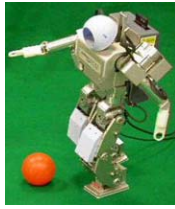<br>(Image courtesy of Prof. CM Chew, National University of Singapore) | Anthropomorphic | - 50 cm tall<br>- Metallic in structure<br>- 25 degrees of freedom: can open/close hands, pan-tilt head<br>- Cameras used for gaze tracking |
| 7 | Infanoid [48] | 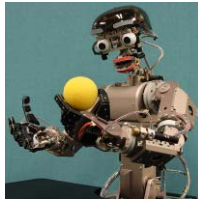<br>(©2000 IEEE. Reprinted, with permission, An Epigenetic Approach to Human-Robot Communication, Kozima H, Zlatev J) | Anthropomorphic | - Upper torso robot<br>- Size of a 4-year old child<br>- Capable of directing gaze and face<br>- Facial expressions with lips and eyebrows<br>- Hand and body gestures<br>- Tracks motion by video processing |
| 8 | IROMEC (Interactive Robotic Social Mediators as Companions) [34,56] | 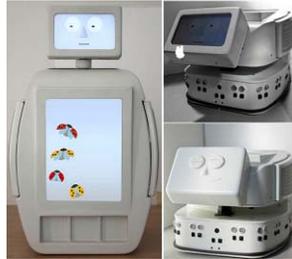<br>©2010 Marti P, Perceiving While Being Perceived, used under a Creative Commons Attribution licence) | Anthropomorphic (vertical mode) and non-biomimetic (horizontal mode) | - Mobile platform<br>- Interfaces for input/output: dynamic screens, buttons, wireless switches<br>- Mask to cover the screen for adaptability<br>- Horizontal: interaction module attached to mobile platform for better mobility<br>- Vertical: interaction module attached to docking station for stability and recharging |
| 9 | Jumbo [35] | 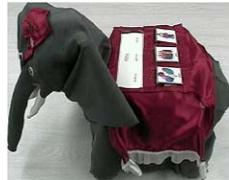<br>(©2003 IEEE. Reprinted, with permission, Characteristics of Mobile Robotic Toys for Children with Pervasive Developmental Disorders, Michaud F, Duquette A, Nadeau I) | Non-anthropomorphic | - Elephant-shaped robot<br>- Moveable head and trunk<br>- Has 3 control buttons to select pictograms located on its back<br>- LEDs to assist child in selection<br>- Modular and adaptable: pictograms can be easily replaced<br>- Robust |
| 10 | KASPAR [41,57] | 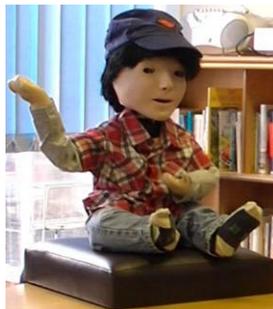<br>©2013 Wood LJ, Dautenhahn K, Rainer A, Robins B, Lehmann H, Syrdal DS, used under a Creative Commons Attribution licence) | Anthropomorphic | - Child-sized, male<br>- Head, arm, and hand movements<br>- Simple facial expressions and gestures<br>- Non-moving torso and legs<br>- Presence of eyelids<br>- Child-like face, plain facial colour, and no facial hair<br>- Adaptable, suited for customizable therapy |
| 11 | Keepon [29,45] | 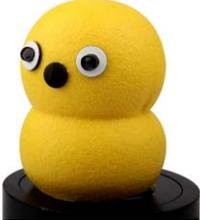<br>(With kind permission from Springer Science+Business Media: Intl J Soc Rob, Keepon, 2009, Kozima H, Michalowski MP, Nakagawa C) | Non-anthropomorphic | • Snowman-like body, yellow in colour and made of silicone rubber<br>• Multi-axis movement: 4 degrees of motion<br>• Built-in microphone inside the nose<br>• Has touch sensors<br>• Cameras inside eyes for video processing<br>• Expresses emotions with body movements: pleasure (side-to-side), excitement (up and down) and fear (vibration) |
| 12 | Kismet [58,59] | 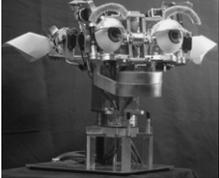<br>(With kind permission from Springer Science+Business Media: Auton Rob, Recognition of Affective Communicative Intent in Robot-Directed Speech, 2002, Breazeal C, Aryananda L) | Anthropomorphic | • Facial features for emotive expression: anger, fatigue, fear, disgust, excitement, happiness, interest, sadness, and surprise<br>• Has a stereo active vision system<br>• Cameras in eyeballs |
| 13 | Labo-I [60-62] | 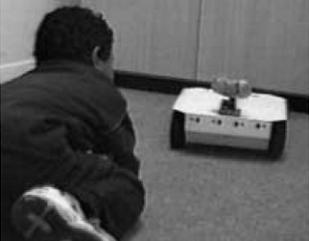<br>(Reprinted with kind permission from John Benjamins Publishing Company, Amsterdam/Philadelphia, Pragmatics & Cognition, Towards Interactive Robots in Autism Therapy, 2004, Dautenhahn K and Werry I) | Non-biomimetic | • Robust and mobile<br>• Medium-sized robot (38 cm long, 30 cm wide and 21 cm high)<br>• Weighs 6.5 kg<br>• Uses infrared and heat sensors<br>• Vehicular structure |
| 14 | Lego Mindstorm NTX [50,63] | 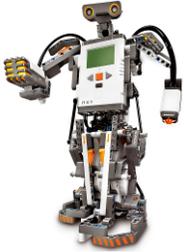<br>(©2013 The LEGO Group. Used with permission) | Anthropomorphic | • Sound and touch sensors for activation<br>• Capable of motion<br>• Modular: made from LEGO bricks<br>• Adaptable and capable of changing features |
| 15 | Maestro [35] | 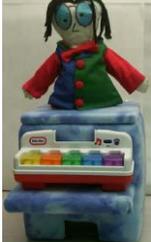<br>(©2003 IEEE. Reprinted, with permission, Characteristics of Mobile Robotic Toys for Children with Pervasive Developmental Disorders, | Anthropomorphic | • Has illuminated keyboard for interaction<br>• Capable of motion<br>• Can play music, vibrate, and dance |

| 16 | Nao [64] | 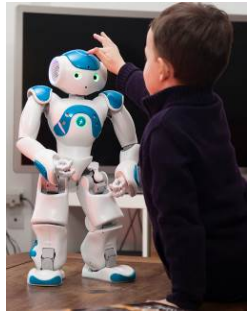<br>(Nao, image courtesy of Aldebaran) | Anthropomorphic | - 50 cm tall<br>- 25 degrees of freedom<br>- Has cameras, microphones, speakers, touch sensors, and LEDs<br>- Capable of speech and touch<br>- Can change eye colour |
|---|---|---|---|---|
| 17 | Paro [65,66] | 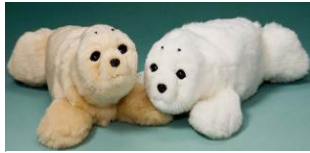<br>(With kind permission from Springer Science+Business Media: Intl J Soc Rob, Investigation on People Living with Seal Robot at Home, 2012, Shibata T, Kawaguchi Y, Wada K) | Non-anthropomorphic | - Seal-like appearance with white fur body<br>- Tactile sensors to detect human contact<br>- Speech recognition and detection of sound source direction<br>- Vertical/horizontal neck movements<br>- Front/back paddle movements<br>- Eyelid movements for facial expressions |
| 18 | Pekee [67] | 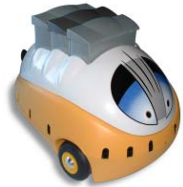<br>(Pekee, image courtesy of Wany Robotics) | Non-biomimetic | - Medium-sized mobile robot (40 cm long, 25 cm wide, 21 cm high)<br>- Automatically records the child's behaviors<br>- Has 3 wheels to move about<br>- Obstruction avoidance while moving<br>- Maximum speed of 6 km/hr |
| 19 | Roball [35,47] | 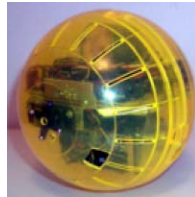<br>(©2003 IEEE. Reprinted, with permission, Characteristics of Mobile Robotic Toys for Children with Pervasive Developmental Disorders, Michaud F, Duquette A, Nadeau I) | Non-biomimetic | - Spherical robot made with a plastic structure<br>- 15 cm in diameter<br>- Weighs about 1.8 kg<br>- Components situated on an internal plateau<br>- Capable of navigating in all directions without getting stuck or falling<br>- Interactions done using vocal messages and movement patterns |
| 20 | Robota [68] | 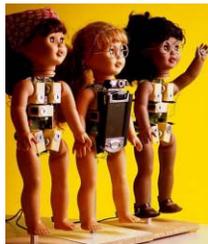<br>(Figure reprinted with permission from Taylor & Francis Ltd: Assistive Technology, Building Robota, 2007, Billard A, Robins B, Nadel, J, Dautenhahn K) | Anthropomorphic | - 45 cm high, 14 cm wide doll<br>- Weighs 1.5 kg<br>- Head rotation<br>- Arm and leg movement (up and down)<br>- Coordinated and individual eye motion<br>- Uses speech processing and video processing for motion tracking<br>- 2 modes: as a puppet and as a dancing toy |

| 21 | Tito [46,47] | 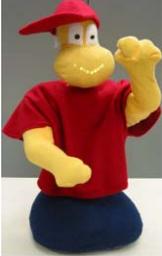 (With kind permission from Springer Science+Business Media: Auton Rob, Exploring the Use of a Mobile Robot as Imitation Agent, 2008, Duquette A, Michaud F, Mercier H) | Anthropomorphic | <ul><li>60 cm tall and is plainly colored (red, yellow, blue)</li><li>Has feet and legs but uses wheels for motion</li><li>Can move arms, head, and mouth</li><li>Can express a few simple emotions</li><li>Uses camera and microphone</li><li>Generates vocal messages</li><li>Different body parts can be illuminated</li><li>Has a vocabulary of 25 words</li></ul> |
|---|---|---|---|---|
| 22 | TREVOR (Triadic Relationship EVOking Robot) [36] | 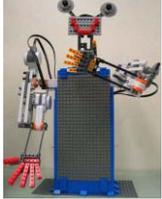 (©2010 IEEE. Reprinted, with permission, Detailed Requirements for Robots in Autism Therapy, Guillian N, Ricks D, Atherton A et al) | Anthropomorphic | <ul><li>Toddler-sized</li><li>Can grasp and move objects</li><li>Modular: made of LEGO bricks</li></ul> |
| 23 | Troy [36,69] | 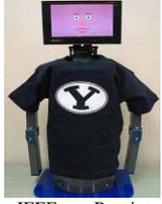 (©2010 IEEE. Reprinted, with permission, Detailed Requirements for Robots in Autism Therapy, Guillian N, Ricks D, Atherton A et al) | Anthropomorphic | <ul><li>Upper-body robot, size of a 4-year old child</li><li>64 cm tall, with 30 cm arm length</li><li>Moveable arms with 4 degrees of freedom</li><li>Weighs 7 kg, has a platform base</li><li>30 pre-programmed actions</li><li>Moveable screen face</li><li>Can move objects</li><li>Can speak and sing</li></ul> |

# Eliciting target behaviours in therapy

The purpose of conducting therapeutic child-robot interaction sessions is to enable the children to overcome their deficiencies and gain a better understanding of the world. These interactions are aimed at improving the children's social skills, emotional awareness, and their communication with the environment and people around them. To achieve these objectives, therapy sessions are composed of activities that can result into positive behaviours from children with autism. This section describes these behaviours and how the robots elicit such behaviours during the therapy sessions.

*Imitation*

Imitation plays a significant role in the transfer of knowledge to the child from an external source. A child not only learns new physical and verbal skills but also explores his/her social environment through imitation. The child also picks up new behavioural traits through this. Imitation activities help to develop a cross-modal mapping mechanism in children [48]. They also improve hand-eye coordination and enable the children to recognize the people around them as their social peers, whose actions they can imitate. These activities are so important that nearly every robot in Table 2 uses imitation in therapy to treat a child with autism. A robot teaches this skill to a child by engaging him/her in simple imitation games [34], which if executed successfully, allow the child to receive sensory rewards and encouragement from the robot.

*Eye contact*

It has been shown that eye gaze can be more useful than verbal communication in order to distinguish between children with autism and a control group of children with moderate learning difficulties [70]. Eye contact and eye gaze form a vital part of social development since they are used to maintain face-to-face interactions. Eye contact serves not only to monitor each other's state of attention and emotion, but also to establish mutual acknowledgement. Since the ability to make and maintain eye contact is naturally deficient in a child with autism, the robot's intervention becomes highly valuable.

*Joint attention*

The act of sharing attentional focus is called joint attention [48]. It is the joint action of two individuals looking at the same target through eye gaze or pointing by means of hand gestures. The ability to maintain focus on a single object is naturally inhibited in children with autism, causing joint-attention activities to be especially difficult for them. During the child-robot interactions, the robot is used to guide the child's attention to a specific object such that the child is easily able to follow the direction of the robot's gaze. As progress is made, the child is able to initiate the act of guiding the robot's attention too and may even extend this behaviour in order to interact with the therapist [45]. This makes joint attention activities very promising in robot-assisted autism therapy.

*Turn-taking*

Children with autism find it extremely difficult to share things and indulge in normal conversations involving taking turns with others [44]. They are generally known to ramble unstoppably about their own obsessive ideas without giving consideration to their partner in the conversation. Activities with therapeutic robots help children develop turn-taking ability, teaching them to wait for responses from a partner before they say or do something. This can be achieved by engaging the child in simple games with the robot, such as the child kicking a ball to the robot, followed by the robot kicking the ball back to the child [51]. Another is the chase-and-avoid game where the child takes turns to first chase a mobile robot and then avoids it as it follows [61]. Using toy robots in group trials also allow the child to learn to wait for his or her turn to play with an object [71].

*Emotion recognition and expression*

It has been observed that children with autism find it very hard to read and interpret facial expressions and body language [1,4]. Interactions with others can involve excessive sensory stimulation, causing severe distress to the child with autism. Child-robot interactions are markedly different since the robots are programmed to show only a small set of basic emotions. These are communicated to the child with simple, indulging activities, hence eliminating any sensory overload [41,47]. Many of the aforementioned robots use simple designs and limited facial expressions to project minimal emotions to the child in therapy.

*Self-initiated interactions*

Another deficiency commonly found in children with autism is their difficulty to ask for things that they need [44]. They find it extremely difficult to initiate interactions themselves. Consequently, they may resort to violent behaviour or tantrums. During the therapy sessions, the clinician encourages the child to ask for toys that the child may want to play with instead of handing the toy over easily. Robots have also been designed to train children for self-initiation [47,53,64,67,68], performing an action only after the child has pressed a button or made a sound. The resulting action of the robot serves as a reward for the child and encourages the child to initiate interactions not just with the robot but outside the therapy session as well.

*Triadic interactions*

A triadic interaction is one that involves a child, a robot, and another companion [44]. The goal of all child-robot interactions is not just for the child to learn the required skills in therapy sessions but to generalize those lessons to the people around them. Eventually, the objective must be to improve the child's social interaction and communication with peers and not with the robot only. Experiments have shown that the presence of robots helps elicit triadic interactions from a child, such as when a child looks to the therapist to share excitement about the robot's actions [45]. Such interactions also instill self-initiation and joint attention skills, and prove to be of great benefit for a child with autism. For some children, simple realizations, such as becoming aware that the robot is being controlled remotely by the therapist, have also evolved into triadic interactions with the therapist [68].

## Child-robot interaction experiments

The dependence of behaviour elicitation on a robot's design features is shown in Table 3. These clinical experiments were conducted to determine the therapeutic value of robots in autism therapy for children.

Different therapy models have been used in the experiments. These include:
  a. Single versus repeated interactions: The experiment may be performed once or a number of times to obtain the results.
  b. Structured versus free-form interactions: Structured interactions involve the presence of a therapist as an active member in the activity. Free-form interactions allow the child to interact with the robot with no interaction from the therapist, unless necessary.
  c. Individual versus group experiments: The experiment may be performed on an individual or multiple individuals.

The experimental data also contains the autism diagnosis of the child in therapy. The child's age may be given both in terms of the chronological age (CA) and mental age (MA). The difference between the two is an indicator of the severity of the disorder [72]. Chronological age is the number of years an individual has lived and mental age is the age at which the child is performing intellectually. These diagnostic details are important to mention since the effectiveness of a therapeutic technique cannot be made independent of the nature of the child's disorder.

This compilation can be helpful because it shows the suitability of a particular robot in terms of the results obtained from the experiments conducted so far. It enables one to determine the robot features that are favourable for eliciting a specific behaviour from a child participant in therapy, hence allowing for the design of a suitable therapeutic robot. Robot designers can derive considerable benefit from this compilation since it allows them to focus on the design process, with knowledge about the therapeutic effectiveness of their design.

Table 3: Experimental data to show the dependence of behaviour elicitation on robot design features

| Targeted behaviour | Robot | Participants | Therapy Model | Method | Findings |
|---|---|---|---|---|---|
| Imitation | Keepon [45] | A 3-year old girl, diagnosed with autism with moderate mental retardation | • Individual interactions<br>• Structured<br>• Repeated interactions (15 sessions over 5 months) | • The child and the caregiver were seated with Keepon on the floor<br>• Unconstrained interaction: no instructions were given<br>• The interaction was allowed to continue until the child lost interest | • The child was attracted to the bobbing and rocking gestures of Keepon<br>• The child initiated an imitation game by mimicking the robot's body movements |
| | FACE [73] | 4 subjects (3 males and 1 female), aged between 7-20 years, all diagnosed with high-functioning autism | • Individual interactions<br>• Structured<br>• One interaction | • The subjects can interact with FACE using a software, through a screen and keyboard/mouse for 20 minutes<br>• The subjects wore a sensorized t-shirt for recording physiological data, since emotions may have been difficult for them to communicate<br>• The subjects were tested for spontaneous imitation of FACE's facial expressions and gestures<br>• The results were evaluated using the Childhood Autism Rating Scale (CARS) (by comparing with CARS rating from the previous therapies) | • 3 out of 4 subjects demonstrated a decrease in imitation score on CARS scale, which signifies an improvement in imitative behaviour |
| | Robota [74] | 4 children with autism, aged between 5-10 years<br>• Child A, aged 5, used only 2-3 words<br>• Child B, aged 6, had limited verbal expression<br>• Child C, aged 10, had no verbal | • Individual interaction<br>• Structured<br>• Repeated interactions (9 trials on an average) | • The robot sat on a table. The therapist operates the robot. The interaction continued as long as the child showed interest (average duration: 3 minutes)<br>• The robot was operated in dancing mode in initial sessions for | • The total number of occurrences of imitation from children yielded a score<br>• The score showed an overall increasing trend, with the highest increment at the endmost trials |
| | | language and severe learning difficulties<br>• Child D, aged 10, had verbal language but little attention and motivation | | familiarization, and then changed to puppet mode later for imitation games<br>• Robot progressively changes interactions from instructed to unconstrained | |
| | KASPAR [41] | A boy, aged 16, lacked focus, was aggressive and could not tolerate other children | • Individual interactions (12 weeks)<br>• Structured<br>• Repeated interaction | • KASPAR was placed on a table, the investigator and the child were seated in front of the robot | • The therapist initiated an imitation game<br>• At first, the subject controlled the robot and the therapist imitated it<br>• Then the therapist controlled the robot and the subject imitated it<br>• The subject observed as the therapist imitated and learnt those actions<br>• The subject showed interest in the robot and explored its features<br>• The subject refused to let go of KASPAR's remote control |
| | | | • Trial by pairs (another child joins the session)<br>• Repeated interactions (12 weeks)<br>• Structured | • KASPAR was placed on a table; the investigator and the child were seated in front of the robot | • After learning the imitation game from the therapist in the earlier session, the subject became less resistant to social play<br>• The subject continued to play the same game with the second child |
| Eye contact | Keepon [45] | Control group: 25 typically developing children in 3 age groups:<br>• 0-1 year ($N$=8, Average 9 months old)<br>• 1-2 years ($N$=8, Average 16.5 months old)<br>• Over 2 years ($N$=9, Average 37.3 months old) | • Individual interactions<br>• Structured | • The child and the caregiver were seated with Keepon on the floor<br>• Unconstrained interaction: no instructions were given<br>• Interaction allowed to continue until the child lost interest<br>• The robot detected the human face and maintained its gaze on the child | The differences in interactions showed differences in ontological understanding between age groups:<br>• 0-1 year: Showed indifference to the robot's attempts to make eye contact<br>• 1-2 years: Showed awareness of the robot's gaze direction<br>• Over 2 years: Engaged in coordinated eye contact on realizing that the robot responded to their actions |
| | | A 3-year old girl, diagnosed with autism with moderate mental retardation | • Individual interactions<br>• Structured<br>• Repeated interactions (15 sessions over 5 months) | | The subject averted the robot's gaze in the first few sessions but started looking into its eyes directly as she grew familiar with it |

| Robot | Participants | Interaction Type | Procedure | Findings |
|---|---|---|---|---|
| Robota [74] | 4 children with autism, aged between 5-10 years<ul><li>Child A, age 5, used only 2-3 words</li><li>Child B, aged 6, had limited verbal expression</li><li>Child C, aged 10, had no verbal language and severe learning difficulties</li><li>Child D, aged 10, had verbal language but little attention and motivation</li></ul> | <ul><li>Individual interactions</li><li>Structured</li><li>Repeated interactions (9 trials on an average for each)</li></ul> | <ul><li>The robot sat on a table. The therapist operates the robot</li><li>Interaction continues as long as child shows interest (average duration: 3 minutes)</li><li>Robot progressively changes interactions from instructed to unconstrained</li></ul> | <ul><li>The total number of occurrences of eye contact from children yielded a score</li><li>The child was given a chair to interact with the robot on day 1 and day 8</li><li>The score was highest for these days but showed an increasing trend from trial 3 onwards</li></ul> |
| Nao [75] | A boy, aged 10 years, had IQ in the average to above average range; high functioning autism | <ul><li>Individual interaction</li><li>Free-form but in the presence of the class teacher for a comforting presence</li><li>Single interaction</li></ul> | <ul><li>Robot executed 5 modules (duration = 14 mins 30 seconds):<br>1. Introductory rapport<br>2. Talking<br>3. Arm movement<br>4. Song play and eye blink<br>5. Song play and arm movement</li><li>The trial was aborted if the child became uncooperative</li><li>The child's behaviour in class was evaluated to compare against the trial data</li><li>A Gilliam Autism Rating Scale 2$^{nd}$ Edition (GARS-2) behaviour score was sheet used to evaluate the child's behavior</li><li>A subset of items were utilized to suit the experiment's needs</li></ul> | <ul><li>Noticeable increase in the child's eye contact was seen: the child avoided the teacher's eye gaze but looked at robot easily, especially when the robot changed eye colour, talked or when arm movements were made</li><li>The child showed improvements in stereotypical behavior, communication and social interaction</li><li>Other children in the same IQ group predicted to show the same response to Nao</li></ul> |
| HOAP-2 [55] | 64 18-month old infants with no known developmental problems; half of them female, the other half male<ul><li>Divided into Groups A and B</li></ul> | <ul><li>Individual interactions</li><li>Structured (Group A) Free-form (Group B)</li></ul> | <ul><li>The robot was placed on a table</li><li>The child was seated in the parents' lap in front of the table</li><li>Group A was exposed to the robot's abilities and they observed the experimenter interacting with the robot (imitation and eye gaze following activities)</li><li>Group B had no previous interaction with the robot</li></ul> | <ul><li>Group A performed better since infants had previously witnessed the robot's actions and abilities</li><li>Group A was able to focus on the robot's eye gaze better than Group B</li></ul> |

| | | | | | |
|---|---|---|---|---|---|
| | | | | - Hidden speakers next to the robot's feet emitted a tone to get the child's attention before the experiment started<br>- The robot moves its head laterally and maintained gaze at an object for 6 seconds before moving again<br>- Scoring: +1 for correct target look, -1 for incorrect look | |
| | KASPAR [41] | A boy diagnosed with severe autism; interacted with family but not with anyone at school. The boy isolated himself | - Individual interactions<br>- Structured<br>- Repeated interaction | - KASPAR was placed on a table, with the investigator and the child seated in front of it<br>- The child engaged in tactile exploration of the robot's features | - The child showed fascination with the robot's eyes and eyelids, and often touched and explored them<br>- This later led to the child touching his own eyes and face and as well as his therapist's eyes and face |
| Joint attention | Keepon [45] | Control group: 25 typically developing children in 3 age groups:<br>- 0-1 year (*N*=8, Average 9 months old)<br>- 1-2 years (*N*=8, Average 16.5 months old)<br>- Over 2 years (*N*=9, Average 37.3 months old) | - Individual interactions<br>- Structured | - The child and the caregiver were seated with Keepon on the floor<br>- Unconstrained interaction: no instructions were given<br>- The experiment was allowed to continue until the child lost interest<br>- Keepon oriented towards the target by directing its gaze (up/down, left/right) to it | The differences in interactions showed differences in ontological understanding between age groups<br>- 0-1year: Showed indifference to the robot's attempted attentive action<br>- 1-2 years: Showed awareness of the robot's attentive actions<br>- Over 2 years: Actively coordinated their attention with the robot |
| | KASPAR [41] | A boy diagnosed with severe autism; interacted with family but not with anyone at school. The boy isolated himself | - Individual interactions<br>- Structured<br>- Repeated interaction | - KASPAR was placed on a table, with the investigator and the child seated in front of it<br>- The child engaged in tactile exploration of the robot's features | - The child turned to the therapist and smiled after an interaction with KASPAR<br>- The robot acted as an object of joint attention: the child gazed and smiled at the therapist in response to KASPAR |
| Emotional attention and expression | Keepon [45] | Control group: 25 typically developing children in 3 age groups:<br>- 0-1 year (*N*=8, Average 9 months old)<br>- 1-2 years (*N*=8, Average 16.5 months old)<br>- Over 2 years (*N*=9, Average 37.3 months old) | - Individual interactions<br>- Structured | - The child and the caregiver were seated with Keepon on the floor<br>- Unconstrained interaction: no instructions were given<br>- The experiment was allowed to continue until the child lost interest | The differences in interactions showed differences in ontological understanding between age groups<br>- 0-1 year: Showed a positive response to emotive actions, e.g. laughing<br>- 1-2 years: Mimicked emotive actions<br>- Over 2 years: Coordinated emotional activities with the robot, such as expression |

| | | | | |
|---|---|---|---|---|
| | | | The robot expressed emotions about its attention target by fixing gaze on it, and rocking left to right, or bobbing up and down | of fondness by soothing its head |
| | A 3-year old girl, diagnosed with autism with moderate mental retardation | - Individual interactions
- Structured
- Repeated interactions (15 sessions over 5 months) | | - The child engaged actively in expressing fondness for the robot by actions such as placing a knitted cap on its head and kissing the robot |
| | A boy, aged 3 years, diagnosed with Asperger's syndrome with mild mental retardation | - Individual interactions
- Structured
- Repeated interactions (15 sessions over 9 months) | | - The child slowly changed from treating Keepon violently to defending it from strangers and other children.
- The subject indulged in conversations with the robot, and asked the robot about its health |

| | | | | | |
|---|---|---|---|---|---|
| | FACE [73] | 4 subjects (3 males and 1 female) between 7 and 20 years old, all diagnosed with high-functioning autism | • Individual interactions<br>• Structured<br>• Single interaction | • The subjects could interact with FACE using a software, through a screen and keyboard/mouse for 20 minutes<br>• The subjects wore a sensorized t-shirt for recording physiological data since emotions may have been difficult for them to communicate<br>• The subjects were tested for focus of attention on FACE<br>• The results were evaluated using the CARS (by comparing with CARS rating from the previous therapies) | • All the subjects demonstrated a decrease (in emotional response score on CARS scale) of between 1 and 0.5 points, which signifies an improvement in emotional behaviour |
| Vocalization | Keepon [45] | Control group: 25 typically developing children in 3 age groups:<br>• 0-1 year (N=8, Average 9 months old)<br>• 1-2 years (N=8, Average 16.5 months old)<br>• Over 2 years (N=9, Average 37.3 months old) | • Individual<br>• Structured interactions | • The child and the caregiver were seated with Keepon on the floor<br>• Unconstrained interaction: no instructions were given<br>• The experiment was allowed to continue until the child lost interest<br>• The child was encouraged to develop a bond with the robot and to communicate with it as a friend during the interactions | • The responses from different age groups were limited by their ability to speak<br>• 0-1 year and 1-2 years group both showed positive reactions such as laughing<br>• Over 2 years group engaged in verbal interaction with robot such as asking it questions |
| | | A 3-year old girl, diagnosed with autism with moderate mental retardation | • Individual interactions<br>• Structured<br>• Repeated interactions (15 sessions over 5 months) | | • The subject vocalized non-words to the robot as if expecting a response |
| Triadic interactions | Keepon [45] | A 3-year old girl, diagnosed with autism with moderate mental retardation | • Individual interactions<br>• Structured<br>• Repeated interactions (39 sessions over 17 months) | • The child and the caregiver were seated with Keepon on the floor<br>• Unconstrained interaction: no instructions were given<br>• The experiment was allowed to continue until the child lost interest | • After initiating an imitation game with Keepon, the child looked at her mother and the therapist to share her wonder and excitement |
| | Labo-1 [61] | 4 boys, aged between 8-12 years | • Individual trials | • The mobile robot plays simple | • The child interacted with the robot and |

| | | | | chasing and following games<br>• The robot moves away from the child if he is too close<br>• The robot moves toward the child if he is too far | cleared obstacles from its path<br>• The child smiled at the experimenter to share his excitement |
|---|---|---|---|---|---|
| | KASPAR [41] | A boy diagnosed with severe autism; interacted with family but not with anyone at school. The boy isolated himself | • Individual interactions<br>• Structured<br>• Repeated interaction | • KASPAR was placed on a table, with the investigator and the child seated in front of it<br>• The child engaged in tactile exploration of the robot's features | • The child turned to the therapist and smiled after an interaction with KASPAR<br>• The interaction with KASPAR caused the child to interact with his therapist<br>• The child shared positive emotions |
| Self-initiated interactions | Keepon [45] | A 3-year old girl, diagnosed with autism with moderate mental retardation | • Individual interactions<br>• Structured<br>• Repeated interactions (15 sessions over 5 months) | • The child and the caregiver were seated with Keepon on the floor<br>• Unconstrained interaction: no instructions were given<br>• The experiment was allowed to continue until the child lost interest | • The child engaged in spontaneous dyadic interactions. For example, the child asked the therapist to put a paper cylinder on the robot's head after observing another child do the same. |
| | | A 3-year old girl, diagnosed with autism with moderate mental retardation | • Individual interactions<br>• Structured<br>• Repeated interactions (39 sessions over 17 months) | | • The child, fascinated by the robot's movements, initiated an imitation game with the robot |
| | Labo-1 [61] | 4 boys, aged between 8-12 years | • Trial by pairs<br>• Free form (teacher and experimenter were present but did not initiate interactions) | • The mobile robot played simple chasing and following games<br>• The robot moved away from the child if he was too close<br>• The robot moved towards the child if he was too far<br>• Both children were required to share the robot and play together | • The children were involved in social, cooperative play<br>• The first child learned how to operate the robot from the experimenter<br>• The second child initiated an interaction with the first and asked him how to operate the robot<br>• The first child explained the procedure |
| | Pekee [67] | 6 typically developing children (i.e. 2 subjects each for type A, B, and C), aged between 5-7 years,<br>• Type A: active/boisterous;<br>• Type B: average<br>• Type C: passive/shy | • Individual interactions<br>• Free form but in the presence of an experimenter<br>• Single interaction | • The interaction was held inside 2 $m^2$ arena, enclosed by 4 walls<br>• The interaction lasted between 1 - 1.5 minutes each<br>• The robot executed wandering or simple obstacle obstruction behaviour<br>• Data were collected from sensor readings and behavioural analysis of video data | • The results were based on the number of times the child interacted with the robot by touching and exploring it<br>• Type A children were the most interactive. They touched, pushed and jumped over the robot<br>• Type B children were curious about the robot. They followed it but did not touch it that much<br>• Type C children were cautious of the robot. They touched it only once and kept their distance from the robot |
| | Troy [69] | 2 boys (Child A, aged 3 years, diagnosed with ASD; Child B, aged 8 years, diagnosed with ASD)<br>• Child B is higher functioning than Child A<br>• both showed social impairment, repetitive behaviour and restricted interests | • Individual interactions<br>• Structured (2 clinicians present)<br>• Repeated trials (16 sessions over 3 months) | • 40 minutes of child-clinician interaction sessions, followed by 10 minutes of child-robot interaction sessions were arranged.<br>• Pre- and post-treatment assessments conducted where the child was to interact with the parent, clinicians and an unfamiliar clinician separately in the absence of the robot | • Child A's initiated engagements increased drastically from 11 before treatment to 120 after treatment<br>• Child B's initiated engagements increased from 48 before treatment to 65 after treatment<br>• Child A's responded engagements increased drastically from 108 before treatment to 488 after treatment<br>• Child B's responded engagements increased from 107 before treatment to 146 after treatment |
| Turn-taking | Keepon [45] | A 3-year old girl, diagnosed with autism with moderate mental retardation | • Individual interactions<br>• Structured<br>• Repeated interactions (39 sessions over 17 months) | • The child and the caregiver were seated with Keepon on the floor<br>• Unconstrained interaction: no instructions were given<br>• The experiment was allowed to continue until the child lost interest | • The child initiated a unidirectional imitation game with Keepon, where Keepon was the imitator. The child observed its response and waited for his turn to make another action. |
| | Labo-1 [61] | 4 boys, aged between 8-12 years | • Individual trials<br>• Free form (teacher and experimenter were present but did not initiate interactions) | • The mobile robot played simple chasing and following games<br>• The robot moved away from the child if he is too close<br>• The robot moved towards the child if he is too far | • The child made an action and waited for the robot's response repeatedly until he understood its behaviour<br>• The child observed the robot's movements and learned how it worked |

| | | | | |
|---|---|---|---|---|
| KASPAR [41] | A boy, aged 16, lacked focus, was aggressive, and could not tolerate other children | <ul><li>Individual interaction</li><li>Structured</li><li>Repeated interactions (12 weeks)</li></ul> | <ul><li>KASPAR was placed on table, with the investigator and the child seated in front of it</li></ul> | <ul><li>The therapist initiated an imitation game</li><li>The child and the therapist took turns to control the robot</li><li>The one who was not controlling the robot imitated its actions</li><li>The same game was played with another child instead of a robot. This promoted social play, imitation and turn-taking</li></ul> |

## Discussion and Conclusion

Why robots? Social robots play several important roles and benefits in the therapy of children with autism. Robots in autism therapy are designed to take up numerous roles, even within the same therapy session. Through games and engaging activities, the robots can interact with the children in order to train them with skills, elicit specific, desirable behaviours, and provide encouragement and positive feedback upon the successful completion of a task. Based on the literatures that we analyzed, we have categorized the roles of these robots into the following:

*As a diagnostic agent.* Autism in children is hard to diagnose before the age of 3 years. Before that age, the higher-level behavioural patterns that need to be examined for diagnostic purposes have not been fully developed [76]. However, early intervention can increase the chances of improvements in the child's behaviour later on. For instance, it has been found that eye-gaze patterns in infants can be used to diagnose autism [77]. Since these patterns develop well before the child has learnt to speak, robotics technology can provide a method for early autism detection. Moreover, a robot's ability to reproduce the same actions from one interaction to another is also important in its role as an autism diagnostic agent. While clinicians, as experienced as they may be, can find it hard to repeat actions during interactions with the children, robots are able to do this by their very nature. This is essential since diagnosing ASD requires checking the child's response to the same actions over a period of time. There have been several robots that were developed toward this direction [78-82].

*As a friendly playmate.* Social robots can participate in enjoyable and engaging play activities with children with autism [34,45,47,51,83]. As opposed to group therapy sessions, one-child-

 

one-robot scenarios allow the robot to direct its entire attention to a single child, with play activities that are personalized to a child's needs and preferences [41,45]. Play forms an integral part of a child's cognitive and social development [84], but children suffering from autism or other developmental disorders are often unable to participate in such activities with other children due to their impaired social interaction and communication abilities [85]. Instead, they choose to play in isolation or in situations involving minimum social interaction. Play activities with social robots encourage safe, enjoyable environments, ensuring that the child is able to interact freely and without fear.

*As a behaviour eliciting agent.* Robots in autism therapy act as behaviour eliciting agents. Important target behaviours include imitation, eye contact, turn-taking and self-initiation, as discussed in the earlier sections. These activities are designed to promote sensory, cognitive, social, emotional, and motor developments [34,45,47,83] in order to improve the deficits caused by the disorder. Some examples of such activities include teaching a child to initiate greetings, to wait for its turn to throw the ball, to follow the robots gaze to an object of interest, and to copy the robots movements as it dances.

*As a social mediator.* A social robot can serve as a mediator between the child and the therapist by training the child with social skills with the purpose of extending the learnt behaviours to the child's social peers. This is achieved with pair or group therapy sessions, where two or more children interact with the same robot together [51,61]. The eventual goal is always to enable the child to generalize the learnt social skills to their social circle, which includes other children, family members, therapists, and teachers [45,61,67].

*As a social actor.* Children with autism, unlike normally developing children, are unable to learn social skills over time since their interaction with their environment is severely inhibited. These children learn context-appropriate behaviour through robots, that is, through indirect experience [48]. These robots serve as actors, enacting suitable behaviours in specific social situations to give the child opportunities to learn. The robot accomplishes this through its predictable but progressively changing actions [61].

*As a personal therapist.* The robot provides personalized therapy for every child, in accordance to the child's preferences, disabilities, and needs [38]. A robot's modular features allow for a customizable appearance depending on the child. For example, the robot's eyes can be removed if a child feels intimidated by them. The robot also increases the complexity of interaction depending on the child's progress to ensure that the child is always learning new skills [35].

Socially interactive robots have emerged and they have evolved into a very important therapy tool for children with autism. There are reasons why these robots are beneficial for autism therapy.

*Robots are less complex than humans*. Because robots are simpler and more predictable, it would be easier for children to follow instructions from a robot than from a human. As they interact with robots, children with autism will not be intimidated with the complexities of verbal and nonverbal communication, thus making the whole communication process much easier [34,35,44,45,49]. Consequently, social robots can be used as tools for diagnosis and intervention. Social robots can also provide support for the parents and clinicians.

*Robots make embodied interactions possible* [63,68]. Due to their physical affordances, interactions involving tactile exploration and physical movements make the robots more engaging and interesting for a child [61]. Robots also naturally support multi-modal interactions, including gestures, speech and touch [86-88]. The ability to touch is missing in therapy through virtual characters and software agents, which gives robots a marked advantage [61]. Ideal therapy sessions involve situations that require the use of the child's speech, sounds, visual cues, and movement, which makes robots more appealing [34,35,38].

*Robots are less intimidating than humans*. Robots not only act as playmates for children, but they can be used as small, colourful toys, ensuring that children can feel at ease during the interaction [35,37,47,65,49]. They can be programmed to adapt their behaviour in accordance to the specific needs of a child with whom it is interacting, hence customizing the therapy for a child [35,38]. While robots are programmed and are thus deterministic, they are more suited to the needs of predictability and repetition of a child with autism [40,46,60,68,49].

As the intent of therapy is to develop skills and competencies in a child that are used in daily-living situations, it is critical that the use of robots in autism therapy include evaluations of the extent to which the target behaviours are demonstrated and sustained in social contexts, even without the presence of robots. Most published studies in the use of social robots in autism do not systematically evaluate the generalizability of the outcomes of robot-mediated therapy in autism. Future works can further look into this.

There is much that can be done to improve the value of the research efforts and technological developments that continue to be directed into this emerging field. The predictors to which autism patients reach more positively must be explored in greater detail. This helps in determining which treatment is more suited for a certain deficiency or characteristic. Real-time collection of important user interaction information specific to the child's preference and progress can also be beneficial. More emphasis must be laid on the change in a child's ability to generalize behaviours to other people as a result of therapy. This is very important since the very purpose of therapy is to facilitate the child's social interaction with other people, not just with the robots.

Future research must build a foundation of theories, models, methods, and tools that can advance the understanding of child-robot interaction and allow experiments to be replicated across research groups. This will require the joint efforts of experts from all the relevant disciplines in order to integrate diverse sources of knowledge and skill.

# Acknowledgments

This work was supported by the National University of Singapore Academic Research Funding Grant No. R-263-000-A21-112.

# Figure and Table Legend

**TABLES**

**Table 1:** Estimated rates of occurrence of autism in different countries

**Table 2:** Compilation of robots employed in the therapy of children with autism

**Table 3:** Experimental data to show the dependence of behaviour elicitation on robot design features

**FIGURES**

**Figure 1:** The triad of impairments in Autism Spectrum Disorder (ASD)

# References

1. American Psychiatric Association. Task Force on DSM-IV (1993) DSM-IV draft criteria. Amer Psychiatric Pub Inc.
2. The National Autistic Society Diagnosis of Autism Spectrum Disorders- a guide for health professionals. http://www.autism.org.uk/working-with/health/patients-with-autism-spectrum-disorders-guidance-for-health-professionals.aspx. Accessed 4 April 2013
3. Cashin A, Barker P (2009) The triad of impairment in autism revisited. Journal of Child and Adolescent Psychiatric Nursing 22 (4):189-193
4. Brookdale Care Specialist Triad of impairments. http://www.brookdalecare.co.uk/what-is-autism#triad. Accessed 23 January 2013
5. Baron-Cohen S, Wheelwright S (1999) 'Obsessions' in children with autism or Asperger syndrome. Content analysis in terms of core domains of cognition. The British Journal of Psychiatry 175 (5):484-490
6. Wall K (2009) Autism and early years practice. Sage Publications Limited,
7. Johnson CP, Myers SM (2007) Identification and evaluation of children with autism spectrum disorders. Pediatrics 120 (5):1183-1215
8. NICHY - National Dissemination Center for Children with Disabilities Autism Spectrum Disorders. http://nichcy.org/disability/specific/autism. Accessed 23 January 2013
9. Grandin T, Scariano M (1996) Emergence: labeled autistic. Warner Books New York,
10. Grandin T (2006) Thinking in pictures: And other reports from my life with autism. Bloomsbury Publishing,
11. Kids Health An Autism Spectrum Disorder. http://kidshealth.org/parent/medical/brain/asperger.html?tracking=P_RelatedArticle. Accessed 23 January 2013
12. Autism Spectrum Disorders Health Center Understanding Autism -- the Basics. http://www.webmd.com/brain/autism/understanding-autism-basics. Accessed 23 January 2013
13. Happé F, Ronald A (2008) The 'fractionable autism triad': A review of evidence from behavioural, genetic, cognitive and neural research. Neuropsychology Review 18 (4):287-304
14. NYU Child Study Center Autistic Disorder and Asperger's Disorder (Pervasive Developmental Disorders): Questions & Answers. http://www.aboutourkids.org/families/disorders_treatments/az_disorder_guide/autistic_disorder_aspergers_disorder_pervasive_deve_1. Accessed 23 January 2013
15. Baio J (2012) Prevalence of autism spectrum disorders: Autism and developmental disabilities monitoring network, 14 Sites, United States, 2008. Morbidity and Mortality Weekly Report. Surveillance Summaries. Volume 61, Number 3. Centers for Disease Control and Prevention
16. Kim YS, Leventhal BL, Koh YJ, Fombonne E, Laska E, Lim EC, Cheon KA, Kim SJ, Kim YK, Lee H, Song DH, Grinker RR (2011) Prevalence of autism spectrum disorders in a total population sample. Am J Psychiatry 168 (9):904-912.
17. Autism Resource Centre (Singapore) (2013) Frequently Asked Questions - On Autism. http://autism.org.sg/main/faq.php. Accessed 23 January 2013
18. Autism Science Foundation (2012) How Common is Autism? http://www.autismsciencefoundation.org/what-is-autism/how-common-is-autism. Accessed 23 January 2013
19. Hughes V (2011) Researchers track down autism rates across the globe. Simons Foundation Autism Research Initiative. http://sfari.org/news-and-opinion/news/2011/researchers-track-down-autism-rates-across-the-globe. Accessed 23 January 2013
20. Al-Farsi YM, Al-Sharbati MM, Al-Farsi OA, Al-Shafaee MS, Brooks DR, Waly MI (2011) Brief report: Prevalence of autistic spectrum disorders in the Sultanate of Oman. Journal of Autism and Developmental Disorders 41 (6):821-825
21. Baron-Cohen S, Scott FJ, Allison C, Williams J, Bolton P, Matthews FE, Brayne C (2009) Prevalence of autism-spectrum conditions: UK school-based population study. The British Journal of Psychiatry 194 (6):500-509
22. Fombonne E, Zakarian R, Bennett A, Meng L, McLean-Heywood D (2006) Pervasive developmental disorders in Montreal, Quebec, Canada: prevalence and links with immunizations. Pediatrics 118 (1):e139-e150
23. Kawamura Y, Takahashi O, Ishii T (2008) Reevaluating the incidence of pervasive developmental disorders: impact of elevated rates of detection through implementation of an integrated system of screening in Toyota, Japan. Psychiatry and clinical neurosciences 62 (2):152-159
24. Oliveira G, Ataíde A, Marques C, Miguel TS, Coutinho AM, Mota‐Vieira L, Goncalves E, Lopes NM, Rodrigues V, Carmona da Mota H (2007) Epidemiology of autism spectrum disorder in Portugal: prevalence, clinical characterization, and medical conditions. Developmental Medicine & Child Neurology 49 (10):726-733

25. Parner ET, Thorsen P, Dixon G, de Klerk N, Leonard H, Nassar N, Bourke J, Bower C, Glasson EJ (2011) A comparison of autism prevalence trends in Denmark and Western Australia. J Autism Dev Disord 41 (12):1601-1608
26. Paula CS, Ribeiro SH, Fombonne E, Mercadante MT (2011) Brief report: Prevalence of pervasive developmental disorder in Brazil: a pilot study. Journal of Autism and Developmental Disorders 41 (12):1738-1742
27. Fong T, Nourbakhsh I, Dautenhahn K (2003) A survey of socially interactive robots. Robotics and Autonomous Systems 42 (3-4)
28. Li H, Cabibihan JJ, Tan YK (2011) Towards an effective design of social robots. International Journal of Social Robotics 3 (4):333-335
29. Kozima H, Michalowski MP, Nakagawa C (2009) Keepon: A playful robot for research, therapy, and entertainment. International Journal of Social Robotics 1 (1):3-18
30. Welch KC, Lahiri U, Warren Z, Sarkar N (2010) An approach to the design of socially acceptable robots for children with autism spectrum disorders. International Journal of Social Robotics 2 (4):391-403
31. Fujimoto I, Matsumoto T, de Silva PRS, Kobayashi M, Higashi M (2011) Mimicking and evaluating human motion to improve the imitation skill of children with autism through a robot. International Journal of Social Robotics 3 (4):349-357
32. Schiavone G, Formica D, Taffoni F, Campolo D, Guglielmelli E, Keller F (2011) Multimodal ecological technology: From child's social behavior assessment to child-robot interaction improvement. International Journal of Social Robotics 3 (1):69-81
33. Dillon G, Underwood J (2012) Computer mediated imaginative storytelling in children with autism. International Journal of Human-Computer Studies 70 (2):169-178.
34. Ferrari E, Robins B, Dautenhahn K Therapeutic and educational objectives in robot assisted play for children with autism. In: Proc. of the 18th IEEE International Symposium on Robot and Human Interactive Communication (RO-MAN) 2009. pp 108-114
35. Michaud F, Duquette A, Nadeau I Characteristics of mobile robotic toys for children with pervasive developmental disorders. In: Proc. of the IEEE International Conference on Systems, Man and Cybernetics, 2003. pp 2938-2943
36. Giullian N, Ricks D, Atherton A, Colton M, Goodrich M, Brinton B Detailed requirements for robots in autism therapy. In: Proc. of the IEEE International Conference on Systems Man and Cybernetics, 2010. pp 2595-2602
37. Woods S (2006) Exploring the design space of robots: Children's perspectives. Interacting with Computers 18 (6):1390-1418.
38. Robins B, Otero N, Ferrari E, Dautenhahn K Eliciting requirements for a robotic toy for children with autism - Results from user panels. In: Proc. of the 16th IEEE International Symposium on Robot and Human interactive Communication (RO-MAN) 2007. pp 101-106
39. Hoa TD, Cabibihan JJ Cute and soft: Baby steps in designing robots for children with autism. In: Proc. of the Workshop at SIGGRAPH Asia, Singapore, 2012.
40. Robins B, Dautenhahn K, Dubowski J (2006) Does appearance matter in the interaction of children with autism with a humanoid robot? Interaction Studies 7 (3):509-542.
41. Robins B, Dautenhahn K, Dickerson P From isolation to communication: a case study evaluation of robot assisted play for children with autism with a minimally expressive humanoid robot. In: Proc. of the 2nd International Conference on Advances in Computer-Human Interactions, 2009. IEEE, pp 205-211
42. Kozima H, Nakagawa C (2006) Interactive robots as facilitators of children's social development. In: Mobile Robots Towards New Applications. pp 269-286
43. Robins B, Dautenhahn K, Boekhorst R, Billard A (2005) Robotic assistants in therapy and education of children with autism: Can a small humanoid robot help encourage social interaction skills? Universal Access in the Information Society 4 (2):105-120
44. Ricks DJ, Colton MB Trends and considerations in robot-assisted autism therapy. In: Proc. of the IEEE International Conference on Robotics and Automation (ICRA), 2010. pp 4354-4359
45. Kozima H, Nakagawa C, Yasuda Y (2007) Children-robot interaction: a pilot study in autism therapy. Progress in Brain Research 164:385
46. Duquette A, Michaud F, Mercier H (2008) Exploring the use of a mobile robot as an imitation agent with children with low-functioning autism. Autonomous Robots 24 (2):147-157
47. Michaud F, Larouche H, Larose F, Salter T, Duquette A, Mercier H, Lauria M Mobile robots engaging children in learning. In: Proc. of the Canadian Medical and Biological Engineering Conference, 2007.
48. Kozima H, Zlatev J An epigenetic approach to human-robot communication. In: Proc. of the 9th IEEE International Workshop on Robot and Human Interactive Communication (RO-MAN) 2000. IEEE, pp 346-351
49. Dautenhahn K (2003) Roles and functions of robots in human society: implications from research in autism therapy. Robotica 21 (4):443-452.

50. Costa S, Resende J, Soares F, Ferreira M, Santos C, Moreira F Applications of simple robots to encourage social receptiveness of adolescents with autism. In: Proc. of the International Conference of the Engineering in Medicine and Biology Society, 2009. IEEE,
51. Costa S, Santos C, Soares F, Ferreira M, Moreira F Promoting interaction amongst autistic adolescents using robots. In: Proc. of the International Conference of the Engineering in Medicine and Biology Society (EMBC), 2010. IEEE, pp 3856-3859
52. Sheridan TB (1992) Telerobotics, automation, and human supervisory control. MIT press,
53. Boccanfuso L, O'Kane JM (2011) CHARLIE: An adaptive robot design with hand and face tracking for use in autism therapy. International Journal of Social Robotics 3 (4):337-347
54. Mazzei D, Billeci L, Armato A, Lazzeri N, Cisternino A, Pioggia G, Igliozzi R, Muratori F, Ahluwalia A, De Rossi D The FACE of autism. In: Proc. of the 18th IEEE International Symposium on Robot and Human Interactive Communication (RO-MAN), 2010. pp 791-796
55. Meltzoff AN, Brooks R, Shon AP, Rao RPN (2010) "Social" robots are psychological agents for infants: A test of gaze following. Neural Networks 23 (8):966-972
56. Marti P (2010) Perceiving While Being Perceived. International Journal of Design 4 (2):27-38
57. Wood LJ, Dautenhahn K, Rainer A, Robins B, Lehmann H, Syrdal DS (2013) Robot-mediated interviews - How effective is a humanoid robot as a tool for interviewing young children? PLoS ONE 8 (3)
58. Breazeal C, Scassellati B A context-dependent attention system for a social robot. In: Proc. of the International Joint Conference on Artificial intelligence, 1999. pp 1146-1151
59. Breazeal C, Aryananda L (2002) Recognition of affective communicative intent in robot-directed speech. Autonomous Robots 12 (1):83-104
60. Dautenhahn K (2007) Socially intelligent robots: dimensions of human-robot interaction. Philos Trans R Soc Lond B Biol Sci 362 (1480):679-704.
61. Dautenhahn K, Werry I, Salter T, Boekhorst R Towards adaptive autonomous robots in autism therapy: varieties of interactions. In: Proc. of the IEEE International Symposium on Computational Intelligence in Robotics and Automation, 2003. pp 577-582
62. Dautenhahn K, Werry I (2004) Towards interactive robotics in autism therapy. Pragmatics & Cognition 12 (1):1-35
63. Costa S, Soares F, Santos C, Ferreira MJ, Moreira F, Pereira AP, Cunha F An approach to promote social and communication behaviors in children with Autism Spectrum Disorders: Robot based intervention. In: Proc. of the IEEE International Symposium on Robot and Human Interactive Communication (RO-MAN), 2011. IEEE, pp 101-106
64. Gillesen J, Barakova E, Huskens B, Feijs L From training to robot behavior: Towards custom scenarios for robotics in training programs for ASD. In: Proc. of the IEEE International Conference on Rehabilitation Robotics (ICORR), 2011. IEEE, pp 1-7
65. Marti P, Pollini A, Rullo A, Shibata T Engaging with artificial pets. In: Proc. of the Conference on European Association of Cognitive Ergonomics, 2005. pp 99-106
66. Shibata T, Kawaguchi Y, Wada K (2012) Investigation on people living with seal robot at home. International Journal of Social Robotics 4 (1):53-63
67. Salter T, Dautenhahn K, Boekhorst R (2006) Learning about natural human–robot interaction styles. Robotics and Autonomous Systems 54 (2):127-134
68. Billard A, Robins B, Nadel J, Dautenhahn K (2007) Building Robota, a mini-humanoid robot for the rehabilitation of children with autism. Assistive Technology 19 (1):37-49
69. Goodrich MA, Colton M, Brinton B, Fujiki M, Alan Atherton J, Robinson L, Ricks D, Hansen Maxfield M, Acerson A (2012) Incorporating a robot into an autism therapy team. IEEE Intelligent Systems 27 (2):52
70. Ruffman T, Garnham W, Rideout P (2001) Social understanding in autism: Eye gaze as a measure of core insights. Journal of Child Psychology and Psychiatry 42 (8):1083-1094
71. Werry I, Dautenhahn K, Ogden B, Harwin W (2001) Can social interaction skills be taught by a social agent? The role of a robotic mediator in autism therapy. Cognitive Technology: Instruments of Mind:57-74
72. DeMyer MK, Barton S, Alpern GD, Kimberlin C, Allen J, Yang E, Steele R (1974) The measured intelligence of autistic children. J Autism Dev Disord 4 (1):42-60.
73. Pioggia G, Sica M, Ferro M, Igliozzi R, Muratori F, Ahluwalia A, De Rossi D Human-robot interaction in autism: FACE, an android-based social therapy. In: Proc. of the 16th IEEE International Symposium on Robot and Human Interactive Communication (RO-MAN) 2007. pp 605-612
74. Robins B, Dautenhahn K, Te Boekhorst R, Billard A (2004) Effects of repeated exposure to a humanoid robot on children with autism. Paper presented at the Cambridge Worskhop on Universal Access and Assistive Technology (CWUAAT),
75. Shamsuddin S, Yussof H, Ismail LI, Mohamed S, Hanapiah FA, Zahari NI (2012) Initial response in HRI-a case study on evaluation of child with Autism Spectrum Disorders interacting with a humanoid robot Nao. Procedia Engineering 41:1448-1455

76. Campolo D, Taffoni F, Schiavone G, Laschi C, Keller F, Guglielmelli E A novel technological approach towards the early diagnosis of neurodevelopmental disorders. In: Proc. of the International Conference of the Engineering in Medicine and Biology Society, 2008. IEEE, pp 4875-4878
77. Scassellati B (2007) How social robots will help us to diagnose, treat, and understand autism. In: Thrun S, Brooks R, Durrant-Whyte H (eds) Robotics Research. Springer Tracts in Advanced Robotics, vol 28. pp 552-563
78. Scassellati B Quantitative metrics of social response for autism diagnosis. In: Proc. of the International Workshop on Robot and Human Interactive Communication (RO-MAN), 2005. pp 585-590
79. Scassellati B, Crick C, Gold K, Kim E, Shic F, Sun G (2006) Social development. IEEE Computational Intelligence Magazine 1 (3):41-47
80. Dickstein-Fischer L, Alexander E, Yan X, Su H, Harrington K, Fischer GS An affordable compact humanoid robot for autism spectrum disorder interventions in children. In: Proc. of the International Conference of the Engineering in Medicine and Biology Society (EMBS), 2011. pp 5319-5322
81. Ranatunga I, Torres NA, Patterson R, Bugnariu N, Stevenson M, Popa DO RoDiCA: A human-robot interaction system for treatment of childhood autism spectrum disorders. In, 2012.
82. Torres NA, Clark N, Ranatunga I, Popa D Implementation of interactive arm playback behaviors of social robot Zeno for autism spectrum disorder therapy. In, 2012.
83. Lehmann H, Iacono I, Robins B, Marti P, Dautenhahn K 'Make it move': playing cause and effect games with a robot companion for children with cognitive disabilities. In: Proc. of the 29th Annual European Conference on Cognitive Ergonomics, 2011. ACM, pp 105-112
84. World Health Organization (1993) The ICD-10 classification of mental and behavioural disorders: diagnostic criteria for research.
85. Besio S (2008) Analysis of critical factors involved in using interactive robots for education and therapy of children with disabilities. Editrice UNI Service,
86. Cabibihan JJ, Wing-Chee S, Pramanik S (2012) Human-recognizable robotic gestures. IEEE Transactions on Autonomous Mental Development 4 (4):305-314
87. Cabibihan JJ, So W-C, Saj S, Zhang Z (2012) Telerobotic pointing gestures shape human spatial cognition. International Journal of Social Robotics 4 (3):263-272
88. Cabibihan JJ, Pattofatto S, Jomaa M, Benallal A, Carrozza MC (2009) Towards humanlike social touch for sociable robotics and prosthetics: Comparisons on the compliance, conformance and hysteresis of synthetic and human fingertip skins. International Journal of Social Robotics 1 (1):29-40

**John-John Cabibihan** was conferred with a PhD in biomedical robotics by the Scuola Superiore Sant'Anna, Pisa, Italy in 2007. He is an Assistant Professor at the Department of Electrical and Computer Engineering of the National University of Singapore. Concurrently, he serves as the Deputy Director of the Social Robotics Laboratory, an Affiliate Faculty Member at the Singapore Institute of Neurotechnologies (SiNAPSE) and Associate Editor of the International Journal of Social Robotics. He was the past Chair of the IEEE Systems, Man and Cybernetics Society (Singapore Chapter; terms: 2011 and 2012). He was the Program Co-Chair of the 2010 International Conference on Social Robotics, Singapore; Program Chair of the 2012 International Conference on Social Robotics at Chengdu, China; and General Chair of the 2013 IEEE International Conference on Cybernetics and Intelligent Systems, Manila, Philippines. He is working on the core technologies towards lifelike touch and gestures for prosthetics and social robotics.

**Hifza Javed** completed her Bachelors in Information and Communication Systems Engineering from the National University of Science and Technology, Pakistan, where she received a gold medal for academic excellence. She graduated from the National University of Singapore with a Master's degree in Electrical Engineering in 2013. Her interests evolved from information security during her undergraduate studies to social robotics research in her postgraduate studies. Her research, in particular, is focused on the use of socially assistive robots for long-term health behaviour change in individuals with chronic conditions.

**Marcelo H. Ang, Jr.** received the B.Sc. degrees (Cum Laude) in Mechanical Engineering and Industrial Management Engineering from the De La Salle University, Manila, Philippines, in 1981; the M.Sc. degree in Mechanical Engineering from the University of Hawaii at Manoa, Honolulu, Hawaii, in 1985; and the M.Sc. and Ph.D. degrees in Electrical Engineering from the University of Rochester, Rochester, New York, in 1986 and 1988, respectively. His work experience includes heading the Technical Training Division of Intel's Assembly and Test Facility in the Philippines, research positions at the East West Center in Hawaii and at the Massachusetts Institute of Technology, and a faculty position as an Assistant Professor of Electrical Engineering at the University of Rochester, New York. In 1989, Dr. Ang joined the Department of Mechanical Engineering of the National University of Singapore, where he is currently an Associate Professor, with a Joint Appointment at the Division of Engineering and Technology Management. He also is the Acting Director of the Advanced Robotics Center, and Deputy Director of the Center for Intelligent Products and Manufacturing Systems. His research interests span the areas of robotics, mechatronics, and applications of intelligent systems methodologies. He teaches both at the graduate and undergraduate levels in the following areas: robotics; creativity and innovation, applied electronics and instrumentation; advanced computing; product design and realization. He is also active in consulting work in these areas. In addition to academic and research activities, he is actively involved in the Singapore Robotic Games as its founding chairman.

**Sharifah Mariam Aljunied** received her training as an Educational Psychologist in London, UK in 1994. She is the Principal Specialist for Educational Psychology in Ministry of Education in Singapore, and is a Chartered Educational Psychologist and Associate Fellow of the British Psychological Society. Dr Aljunied has made significant contributions to raising the quality of education on a national level for students with special needs in Singapore. Her diverse research work encompasses the development of effective strategies for the assessment and support of students with learning differences, including students with autism and dyslexia. With over 15 years of extensive experience working with mainstream schools in Singapore, Dr Mariam Aljunied has been instrumental in providing diagnosis and advice as well as implementing professional development for practitioners and teachers in the area of special needs.